\documentclass{article} 
\usepackage{iclr2025_conference,times}

\usepackage{amsmath,amsfonts,bm}

\def\eqref#1{equation~\ref{#1}}

\def\1{\bm{1}}

\DeclareMathAlphabet{\mathsfit}{\encodingdefault}{\sfdefault}{m}{sl}
\SetMathAlphabet{\mathsfit}{bold}{\encodingdefault}{\sfdefault}{bx}{n}

\usepackage{hyperref}
\usepackage{url}

\usepackage{graphicx}
\usepackage{amsfonts}
\usepackage{amsmath}
\usepackage{siunitx}
\usepackage{amssymb}
\usepackage{pifont}
\usepackage{nicefrac}
\usepackage{multirow}
\usepackage{subcaption}
\usepackage[font=small]{caption}
\usepackage{soul}
\usepackage{booktabs}
\usepackage{xcolor}
\usepackage{comment}
\usepackage{sidecap}
\usepackage{xspace}
\usepackage[most]{tcolorbox}
\usepackage{enumitem}
\usepackage{ulem}
\usepackage{makecell}
\usepackage[textsize=tiny,textwidth=24mm]{todonotes}

\usepackage[utf8]{inputenc}
\usepackage[T1]{fontenc}
\usepackage{CJKutf8}

\definecolor{bblue}{HTML}{4F81BD}
\definecolor{rred}{HTML}{c4260b}
\definecolor{ggreen}{HTML}{098c1f}
\definecolor{ppurple}{HTML}{9F4C7C}
\definecolor{oorange}{HTML}{F79646}

\newcommand{\redllm}{{RedLLM}\xspace}
\newcommand{\deconlyllm}{{DecLLM}\xspace}

\usepackage{tablefootnote}

\title{Encoder-Decoder or Decoder-Only? \\ Revisiting Encoder-Decoder Large Language Model}

\author{Biao Zhang\thanks{Correspondence to: \texttt{biaojiaxing@google.com}.},~ Yong Cheng,~ Siamak Shakeri,~ Xinyi Wang\thanks{Work done while at Google.},~ Min Ma,~ Orhan Firat \\
Google DeepMind
}

\iclrfinalcopy 
\begin{document}

\maketitle

\begin{abstract}

Recent large language model (LLM) research has undergone an architectural shift from encoder-decoder modeling to nowadays the dominant decoder-only modeling. This rapid transition, however, comes without a rigorous comparative analysis especially \textit{from the scaling perspective}, raising concerns that the potential of encoder-decoder models may have been overlooked.
To fill this gap, we revisit encoder-decoder LLM (\redllm), enhancing it with recent recipes from decoder-only LLM (\deconlyllm). We conduct a comprehensive comparison between \redllm, pretrained with prefix language modeling (LM), and \deconlyllm, pretrained with causal LM, at different model scales, ranging from $\sim$150M to $\sim$8B.
Using RedPajama V1 (1.6T tokens) for pretraining and FLAN for instruction tuning, our experiments show that \redllm produces compelling scaling properties and surprisingly strong performance. While \deconlyllm is overall more compute-optimal during pretraining, \redllm demonstrates comparable scaling and context length extrapolation capabilities. After instruction tuning, \redllm achieves comparable and even better results on various downstream tasks while enjoying substantially better inference efficiency.  
We hope our findings could inspire more efforts on re-examining \redllm, unlocking its potential for developing powerful and efficient LLMs.

\end{abstract}

\section{Introduction}

A crucial lesson from the past decade for modeling is to design scalable and universal architectures being capable of handling different tasks (modalities) and automatically learning from massive unlabeled data, as remarked by the tremendous success of large language models (LLMs) in various research areas, particularly natural language processing~\citep{openai2023gpt4,team2024gemini}. Different architectures often introduce distinct inductive biases, ultimately affecting a model's learnability, scalability, and generalization ability~\citep{narang2021zhenzhong}. In language modeling (LM), two architectures stood out from the literature: \textit{encoder-decoder} and \textit{decoder-only}. The former utilizes an encoder to process the input and a separate decoder to generate the target~\citep{t5_paper}, while the latter jointly models input understanding and target generation via a single module~\citep{Radford2019LanguageMA}. Both architectures present unique strengths and weaknesses, leading to hot debates regarding their respective superiority~\citep{pmlr-v162-zhang22h,pmlr-v162-wang22u,fu2023decoder}.

However, there has been a recent shift towards favoring decoder-only architectures for LLMs, featured by popular LLMs such as LLaMA~\citep{touvron2023llama}, Gemma~\citep{team2024gemma} and Mistral~\citep{Jiang2023Mistral7}, with very few exceptions~\citep{li2023openba}. Such a shift from our understanding mostly results from the success of GPT models~\citep{Radford2019LanguageMA,brown2020language}, rather than the incapability of the encoder-decoder architecture. For example, when provided with comparable compute (by setting encoder-decoder models with \textit{twice} the parameter count of decoder-only models), \citet{pmlr-v162-wang22u} showed that the encoder-decoder model significantly outperformed its decoder-only counterpart after instruction tuning; \citet{tay2022ul2} highlighted the importance of pretraining objective and proposed UL2, which, combined with encoder-decoder modeling, achieves superior pretraining and finetuning performance across various tasks. These findings present direct evidence about the high potential of encoder-decoder modeling, challenging the recent paradigm shift. Unfortunately, the above studies evaluate architectures solely on an ``apple-to-apple'' basis without accounting for their scaling property~\citep{kaplan2020scaling} -- a crucial factor in modern LLMs, which is exactly the focus of our study.

\begin{figure}[!t]
    \small
    \centering
    \begin{minipage}{0.45\textwidth}
        \centering
        \subcaptionbox{\label{fig:redllm_vis}\redllm}{
            \includegraphics[width=\textwidth]{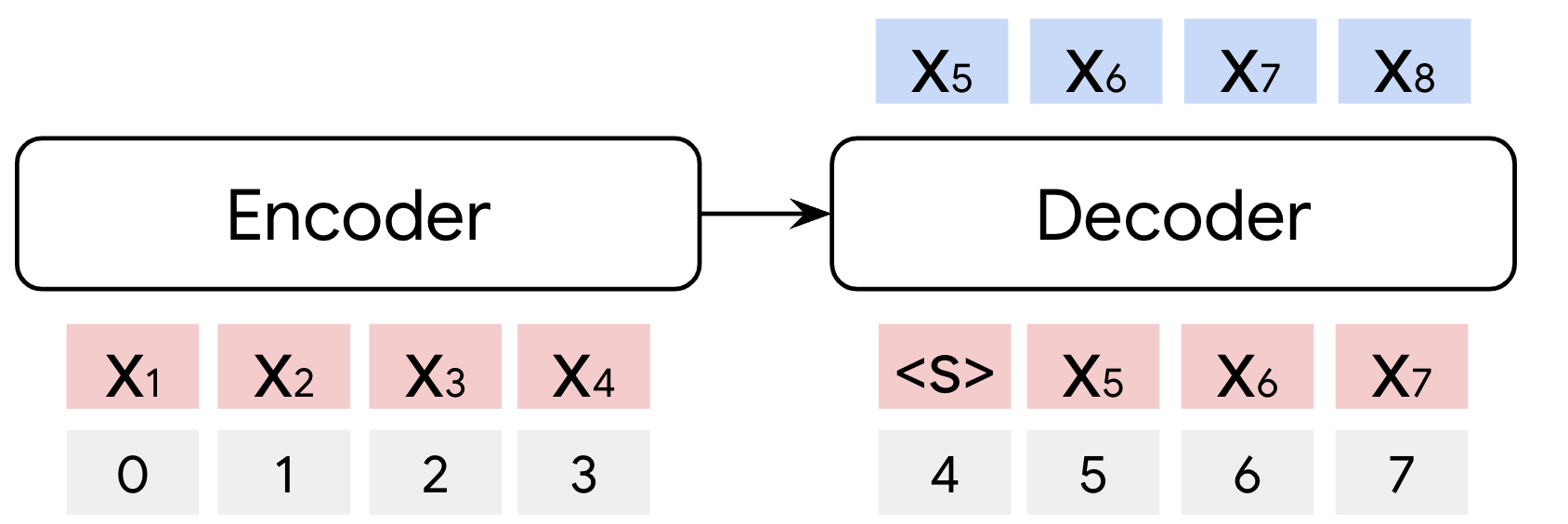}
        }
        \subcaptionbox{\label{fig:decllm_vis}\deconlyllm}{
            \includegraphics[width=\textwidth]{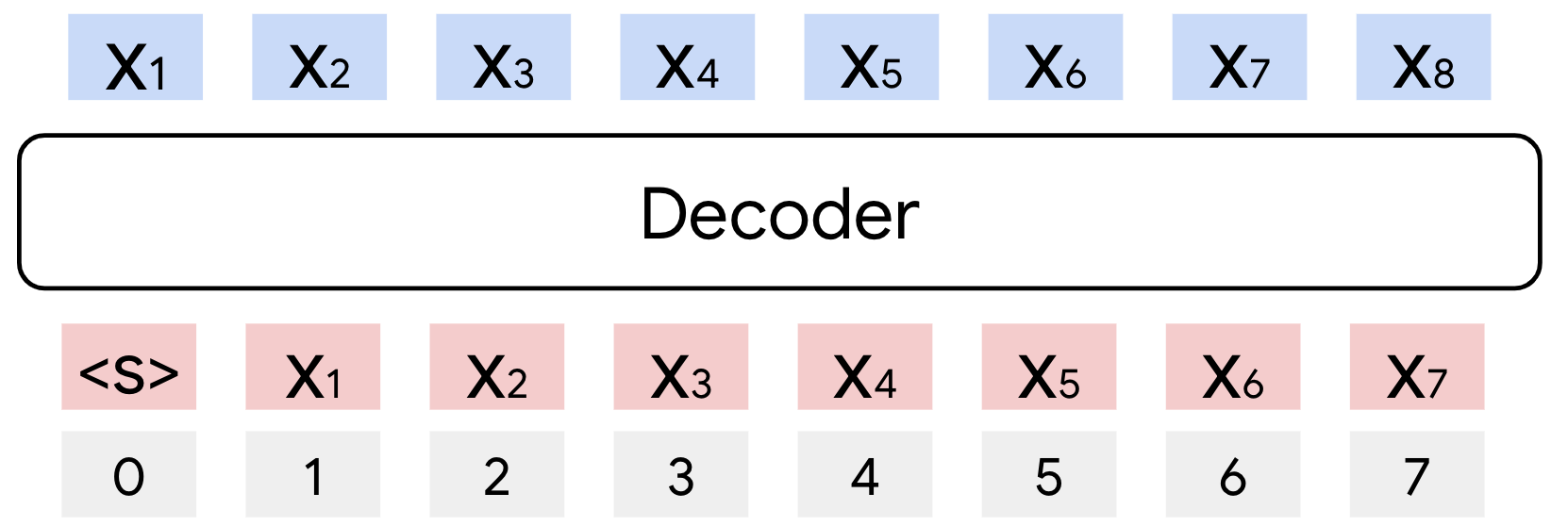}
        }
    \end{minipage}
    ~
    \begin{minipage}{0.53\textwidth}
    \small
    \centering
    \setlength{\tabcolsep}{2pt}
    \subcaptionbox{\label{fig:model_spec}Model specification}{
    \begin{tabular}{lcc}
    \toprule
        & \deconlyllm & \redllm \\
    \midrule
    Attention & \multicolumn{2}{c}{Multi-Head Dot-Product Attention} \\
    FFN Activation & \multicolumn{2}{c}{SwiGLU} \\
    LayerNorm & \multicolumn{2}{c}{RMSNorm (Pre-Normalization)} \\
    Position \\
    \quad Modeling & \multicolumn{2}{c}{Rotary Embedding} \\
    \quad Type & \multicolumn{2}{c}{Continuous Position} \\
    Embeddings & \multicolumn{2}{c}{All Tied} \\
    \midrule
    Extra Norm &  Q, K, V & Q, K, V, Attn Output \\
    Rotary Usage & Self-Attention & Self\&Cross-Attention \\
    Loss & Causal LM & Prefix LM \\
    \bottomrule
    \end{tabular}}
    \end{minipage}
    \caption{\label{fig:overview}Overview of model architecture and specification for \redllm and \deconlyllm. We use red, blue and gray to denote input tokens, output tokens, and positions, respectively. For \redllm, we apply rotary embedding to all attentions (encoder/decoder self-attention and cross-attention) with continuous positions, i.e. decoder position continues from the last one in the encoder. We adopt prefix language modeling for pretraining, and apply layer normalization to query (Q), key (K), value (V), and attention output to improve stabilization.}
\end{figure}

In this paper, we fill this gap by revisiting encoder-decoder LLM (\redllm) and comparing it with decoder-only LLM (\deconlyllm) from the scaling perspective. We adapt recent modeling recipes from \deconlyllm to enhance \redllm, especially the rotary positional embedding~\citep{su2024roformer} with continuous positions, and pretrain \redllm with the prefix LM objective due to its simplicity and efficiency in leveraging compute. Figure \ref{fig:overview} shows the overview. We investigate the scaling properties of \redllm and \deconlyllm by comparing their pretraining and finetuning performance on a range of tasks at various model scales, from approximately 150M to 8B parameters.

We conduct experiments by first pretraining on RedPajama V1~\citep{together2023redpajama} \textit{from scratch} for roughly 1.6T tokens and then finetuning on FLAN~\citep{pmlr-v202-longpre23a} for instruction following. We perform scaling analysis on samples from RedPajama (\textit{in-domain}) and Paloma~\citep[\textit{out-of-domain,}][]{magnusson2023paloma}, and evaluate the models' zero- and few-shot capability on 13 downstream tasks. Our analysis reveals a trade-off between model quality and efficiency, with \redllm and \deconlyllm each excelling in different areas. Main findings are summarized below:
\begin{itemize}
    \item \redllm and \deconlyllm show similar scaling exponents, while \deconlyllm almost dominates the compute-optimal frontier.
    \item During pretraining, \redllm performs badly at zero-shot learning; its few-shot capability scales slightly with model sizes but still lags far behind \deconlyllm.
    \item After instruction tuning, \redllm achieves comparable zero- and few-shot performance to \deconlyllm across scales while enjoying significantly better inference efficiency.
    \item At finetuning, \redllm benefits from the bidirectional attention in its encoder; adapting \deconlyllm with this structure also yields significant improvements. Still, \redllm provides the overall best quality-efficiency trade-off.
    \item \redllm also shows promising context-length extrapolation capability.
\end{itemize}

\section{Background}

Past few years have witnessed the significant progress achieved via large-scale pretraining. By learning from a large amount of unlabeled (weakly-labeled) data, neural networks could acquire massive world knowledge that is transferable to improve downstream tasks. Depending on what to transfer, previous efforts can be roughly classified into two categories: \textit{representation learning} and \textit{generative learning}. The former focuses on pretraining reusable modules that deliver high-quality representations being adaptable to different tasks via task-specific heads~\citep{simonyan2014deep,devlin-etal-2019-bert,liu2019roberta,baevski2020wav2vec,dosovitskiy2020image}. In contrast, the latter unifies pretraining and downstream tasks to maximize the transfer by formulating all downstream tasks as generation tasks, allowing the model to be pretrained entirely in a generative manner~\citep{t5_paper,xue-etal-2022-byt5,lewis-etal-2020-bart,zhang2022opt,scao2022bloom,anil2023palm}. In this study, we mainly explore generative learning, as represented by LLMs.

There are two crucial factors affecting the effectiveness of LLMs: \textit{model architecture} and \textit{pretraining objective}. Model architecture determines the layout and structure of a neural network, directly shaping its generalization. In the literature, two architectures have been widely investigated: {encoder-decoder} and {decoder-only}, both with Transformer as the backbone~\citep{transformer}. The encoder-decoder architecture decomposes the input-target mapping into three separate modules -- encoder self-attention, decoder self-attention and cross-attention for modeling input-input, target-target, target-input dependency, respectively. This decomposition often improves sample and inference efficiency, making it a preferred choice for several LLMs, such as T5~\citep{t5_paper}, BART~\citep{lewis-etal-2020-bart} and OpenBA~\citep{li2023openba}. Instead, the decoder-only architecture relies on a single decoder self-attention module to handle all dependencies, unifying understanding and generation and simplifying the learning. Studies comparing both architectures are many~\citep{fu2023decoder,t5_paper,pmlr-v162-wang22u}, but none of them explored the scaling landscape except~\citep{pmlr-v162-zhang22h} which nevertheless focuses on machine translation exclusively.

Pretraining objective defines what to learn and how to represent information, influencing the model's knowledge acquisition and learning efficiency. Popular choices include span corruption that masks out random tokens from the input to form the target~\citep{t5_paper}, causal LM that simply predicts the next token (i.e., $p(x_1, \ldots, x_T) = \prod_{t=1}^{T} p(x_t|x_{<t})$, $T$ is sequence length)~\citep{Radford2018ImprovingLU}, and prefix LM that follows causal LM but is conditioned on a prefix (i.e., $p(x_k, \ldots, x_T|x_{<k}) = \prod_{t=k}^{T} p(x_t | x_{<t})$, $k$ is the prefix length)~\citep{NEURIPS2019_c20bb2d9}. \citet{tay2022ul2} found that different objectives result in models with different capabilities and limitations, and that mixing them properly produces the best performance albeit at the cost of high implementation complexity and potentially reduced training efficiency~\citep{li2023openba}. We mainly employ the causal LM and prefix LM objective for the pretraining for their simplicity, leaving the exploration of other alternatives to the future.

The performance of pretrained models measured by perplexity is empirically observed following a power-law function of training data size and model size~\citep{kaplan2020scaling,pmlr-v162-bansal22b,pmlr-v202-fernandes23a}, allowing us to analyze compute-optimal scaling for LLM development~\citep{hoffmann2022training}. In addition, pretrained LLMs with sufficient model size often exhibits promising zero/few-shot capability, i.e., solving a task based on a few to no input-output examplers~\citep{brown2020language,team2024gemini}. This task solving ability can be further elicited through instruction tuning, a procedure finetuning LLM on massive downstream tasks~\citep{sanh2022multitask,pmlr-v202-longpre23a,ouyang2022training}. We follow these practices comparing \redllm and \deconlyllm under scaling, and examine their pretraining and finetuning performance with zero/few-shot evaluation.

\section{\redllm: Revisiting Encoder-Decoder LLM}

We adopt variants of Transformer~\citep{transformer} as the LLM backbone, and incorporate several recent techniques from \deconlyllm to improve the encoder-decoder modeling. Table \ref{fig:model_spec} summarizes the model specifications.

For \deconlyllm, we employ the multi-head dot-product attention~\citep{transformer} for the self-attention, SwiGLU~\citep{shazeer2020glu} as the activation for the feed-forward layer, RMSNorm~\citep{zhang2019root} for normalization with the pre-normalization structure, rotary embedding~\citep{su2024roformer} for positional encoding, and tied word embeddings. We apply extra normalization layers to query ($\mathbf{Q} \in \mathbb{R}^{T\times d_h}$), key ($\mathbf{K} \in \mathbb{R}^{T\times d_h}$), and value ($\mathbf{V} \in \mathbb{R}^{T\times d_h}$) vectors over the head dimension ($d_h$) to enhance the training stability
\begin{equation}
    \text{Attn}_{\text{\deconlyllm}} = \text{Softmax}\left(\frac{\text{LN}(\mathbf{Q}){\text{LN}(\mathbf{K})}^T}{\sqrt{d_h}}\right) \text{LN}(\mathbf{V}),
\end{equation}
where $\text{LN}(\cdot)$ denotes parameter-free RMSNorm. Note because of the normalization and the scaling factor $\nicefrac{1}{\sqrt{d_h}}$, the attention logits are bounded into $[-\sqrt{d_h}, \sqrt{d_h}]$, thus avoiding logits explosion.

\redllm follows \deconlyllm. We apply the rotary embedding to all attentions, covering encoder self-attention, decoder self- and cross-attention. As shown in Figure \ref{fig:redllm_vis}, we employ continuous position following \deconlyllm so positional information flows continuously from the encoder's end to the decoder's beginning. We tie all word embeddings, including encoder word embedding, decoder word embedding and output embedding. We note in experiments that \redllm suffers more from training instability. To alleviate this problem, we add another normalization layer to attention output
\begin{equation}
    \text{Attn}_{\text{\redllm}} = \text{LN}\left(\text{Attn}_{\text{\deconlyllm}}\right).
\end{equation}

We pretrain \deconlyllm with the causal LM objective as this objective could fully leverage \deconlyllm's capacity and aligns with the current standard. We use the prefix LM objective for \redllm, instead. While the above specifications are functional, we acknowledge that further architectural/objective optimization is very possible for both models, which we leave to the future.

\begin{table}[t]
\centering
\setlength{\tabcolsep}{2pt}
\small
\begin{subtable}[b]{0.45\textwidth}
\centering
\small
\begin{subtable}[b]{\textwidth}
\begin{tabular}{lrrrrrr}
\toprule
Model Size & $d$ & $d_{ffn}$ & $h$ & $d_h$ & $L_{dec}$ & $L_{red}$ \\
\midrule
150M & 1024 & 4096 & 8 & 128 & 8 & 3/3 \\
1B & 2048 & 8192 & 16 & 128 & 16 & 7/7 \\
2B & 2560 & 10240 & 20 & 128 & 20 & 9/9 \\
4B & 3072 & 12288 & 24 & 128 & 24 & 10/10 \\
8B & 4096 & 16384 & 32 & 128 & 32 & 14/14 \\
\bottomrule
\end{tabular}
\caption{\label{tab:model_parameters} Configurations for different-sized LLMs.}
\end{subtable}
\begin{subtable}[b]{\textwidth}
\vspace{6pt}
\centering
\small
\begin{tabular}{lrrrr}
\toprule
 & \multicolumn{2}{c}{Training Flops} & \multicolumn{2}{c}{\#Params} \\
 \cmidrule(lr){2-5}
 & Dec & Red & Dec & Red \\
\midrule
RedPajama & 0.20 & 0.24 & 0.17 & 0.18 \\
Paloma & 0.24 & 0.27 & 0.20 & 0.20 \\
\bottomrule
\end{tabular}
\caption{\label{tab:scaling_exponents} Fitted scaling exponents.}
\end{subtable}
\end{subtable}
\begin{subtable}[b]{0.50\textwidth}
\centering
\small
\begin{tabular}{lcc}
\toprule
& Pretraining & Finetuning \\
\midrule
Vocabulary & \multicolumn{2}{c}{32768} \\
Dataset & RedPajama V1 & FLAN \\
Steps & 400K & 190K \\
Batch Size & 2048 & 1024 \\
\multirow{2}{*}{Sequence Length} & \deconlyllm: 2048  & \multirow{2}{*}{2048/512} \\
                                 & \redllm: 1024/1024 \\
Optimizer & \multicolumn{2}{c}{Adafactor(decay=0.8)} \\
\multirow{2}{*}{LR Schedule} & 2k-step warmup to 0.01 & \multirow{2}{*}{fixed, 0.001} \\
                             & + cosine decay by 0.1 \\
Gradient Clip & \multicolumn{2}{c}{1.0} \\
Dropout & 0.0 & 0.05 \\
Z-Loss & \multicolumn{2}{c}{0.0001} \\
Precision & \multicolumn{2}{c}{bfloat16} \\
\bottomrule
\end{tabular}
\caption{\label{tab:hyperparameters} Hyperparameters for pretraining and finetuning.}
\end{subtable}

\caption{\label{tab:setup} Settings and scaling. $d, d_{ffn}, d_h$: model, feed-forward, and head dimension, respectively. $h$: number of heads. $L_{dec}, L_{red}$: number of layers for \deconlyllm and \redllm, respectively. Note we adopt a balanced \redllm architecture with equal number of encoder and decoder layers.``B'': billion. ``K'': thousand. ``Dec/Red'': \deconlyllm/\redllm.}
\end{table}

\section{Setup}

We train a set of \redllm{s} and \deconlyllm{s} with model size ranging from about 150M to 8B using T5X~\citep{roberts2023scaling}. Table \ref{tab:model_parameters} lists the model configurations, and Table \ref{tab:hyperparameters} shows the pretraining and finetuning hyperparameters.

\paragraph{Pretraining}

We use RedPajama V1~\citep{together2023redpajama} for pretraining, which is an open-source reproduction of the LLaMA training corpus~\citep{touvron2023llama}. We adopt a subword tokenizer of 32K vocabulary for tokenization.  To train \deconlyllm, we concatenate the documents (separated by tag ``[EOD]'') and then split them into sequences of 2048 tokens (i.e., $T=2048$) without overlapping. We further split each sequence in the middle, resulting in 1024 input and 1024 target tokens (i.e., $k=1024$), for training \redllm. We use Adafactor without factorization~\citep{shazeer2018adafactor} for optimization. All LLMs are trained for 400K steps, totaling roughly 1.6 epochs and 1.6T tokens. Note the effective target token count for \redllm is 0.8T, \textit{half} the size used for \deconlyllm due to the architectural difference.

\paragraph{Finetuning}

We use the FLAN collection~\citep{pmlr-v202-longpre23a} for instruction tuning, which consists of 1800+ tasks with diverse prompts. We set the maximum input/output tokens to 2048/512, and finetune models on the default data mixture (without packing). We tune all model parameters~\citep{zhang2024when}, and compute the log-likelihood loss \textit{only} on the target output. Depending on the model, its size and resource availability, we use up to 256/64 TPU v5p chips for pretraining/finetuning, taking up to 1 month each.

\paragraph{Evaluation}

We perform scaling-law analysis for pretrained models using perplexity (PPL) on RedPajama and Paloma~\citep{magnusson2023paloma}. RedPajama is a random subset of RedPajama V1, including 79K documents sampled from 7 sources (Arxiv, Books, C4, CC, Github, Stackexchange and Wiki). We regard RedPajama as an \textit{in-domain} eval set. Paloma, instead, is independently curated for perplexity analysis, including data from 16 sources and 500+ domains. We argue its independent construction and broad domain coverage make it sufficiently distinct from our pretraining data, thus we consider evaluations using Paloma to be \textit{out-of-domain}. We report averaged PPL over different sources for the evaluation. By default, we use the final checkpoint for pretraining analysis.

We also evaluate zero/few-shot performance of pretrained and finetuned models. We follow~\citet{pmlr-v202-longpre23a} and extend the evaluation to 13 tasks: BoolQ~\citep{clark2019boolq}, HellaSwag~\citep{zellers2019hellaswag}, RTE~\citep{bentivogli2009fifth}, ARC easy and challenge~\citep{allenai:arc}, ANLI R1, R2 and R3~\citep{nie-etal-2020-adversarial}, CommonsenseQA~\citep{talmor-etal-2019-commonsenseqa}, StrategyQA~\citep{geva2021did}, TriviaQA~\citep{2017arXivtriviaqa}, UnifiedQA~\citep{2020unifiedqa}, MMLU~\citep[CoT][]{hendrycks2021measuring}, GSM8K~\citep{cobbe2021gsm8k}, BIG-Bench Hard~\citep[BBH CoT,][]{srivastava2023beyond}, and WMT~\citep[English$\leftrightarrow$German, English$\leftrightarrow$Chinese,][]{kocmi-etal-2022-findings}. We report accuracy except WMT for which we use ChrF. We employ 5 shots for few-shot evaluation except BBH (3 shots), GSM8K (8 shots), CommonsenseQA (7 shots), and StrategyQA (6 shots). We adopt greedy decoding for all models. We calculate averaged score over all tasks for model comparison, and report the best results for finetuning.

\section{Pretraining Results}

\begin{figure}[t]
    \small
    \centering
    \includegraphics[width=0.7\textwidth]{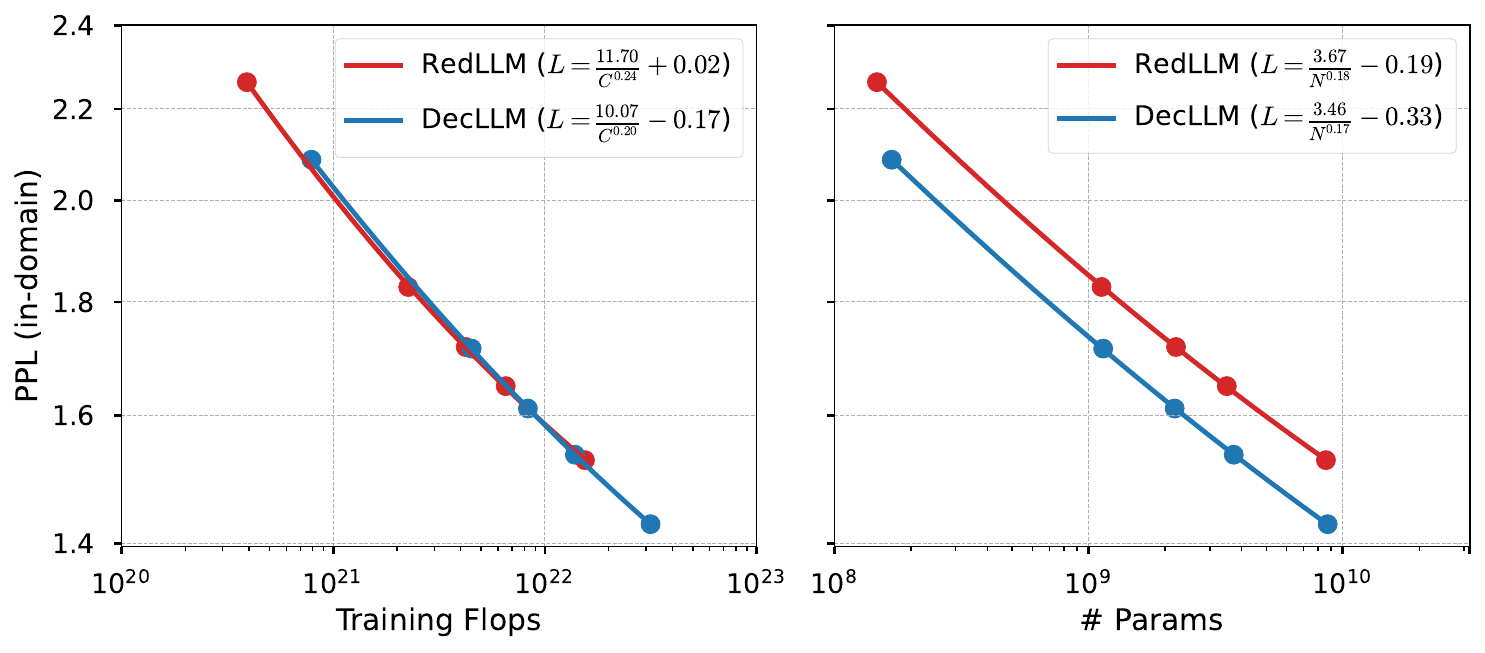}
    \caption{\label{fig:scaling_law_indomain} Fitted scaling law on in-domain dataset (RedPajama) for \redllm and \deconlyllm. Left: training Flops ($C$); Right: model parameters ($N$). To ensure fair comparison, we evaluate PPL using a prefix LM approach, where we utilize the first 1024 tokens as a prefix and compute log-likelihood on the subsequent 1024 tokens.}
\end{figure}

\paragraph{\redllm and \deconlyllm scale at similar rates when increasing the compute (training Flops) and model parameters.}

The substantial differences in architecture and pretraining objective make it challenging to find a common ground comparing \redllm and \deconlyllm. Doubling model parameters for \redllm as in~\citep{pmlr-v162-wang22u,tay2022ul2} is arguably unfair. We instead resort to scaling law~\citep{kaplan2020scaling} and compare them under two factors: model parameters and training Flops. Figure \ref{fig:scaling_law_indomain} and \ref{fig:scaling_law_outdomain} show the results; Table \ref{tab:scaling_exponents} summarizes the scaling exponents. We also show the pretraining loss curve in Figure \ref{fig:training_loss}.

Interestingly, \redllm shows comparable scaling capability to \deconlyllm regardless of evaluation domains, as demonstrated by similar scaling exponents. \deconlyllm is more parameter efficient, outperforming \redllm by a consistent gap under the same amount of parameters. However, this comes at the cost of computational inefficiency, as it requires approximately double the Flops to train under similar conditions. When shifting the comparison basis to compute, the quality gap almost disappears  where both scaling curves nearly overlap. 

\begin{figure}[t]
    \small
    \centering
    \includegraphics[width=0.40\textwidth]{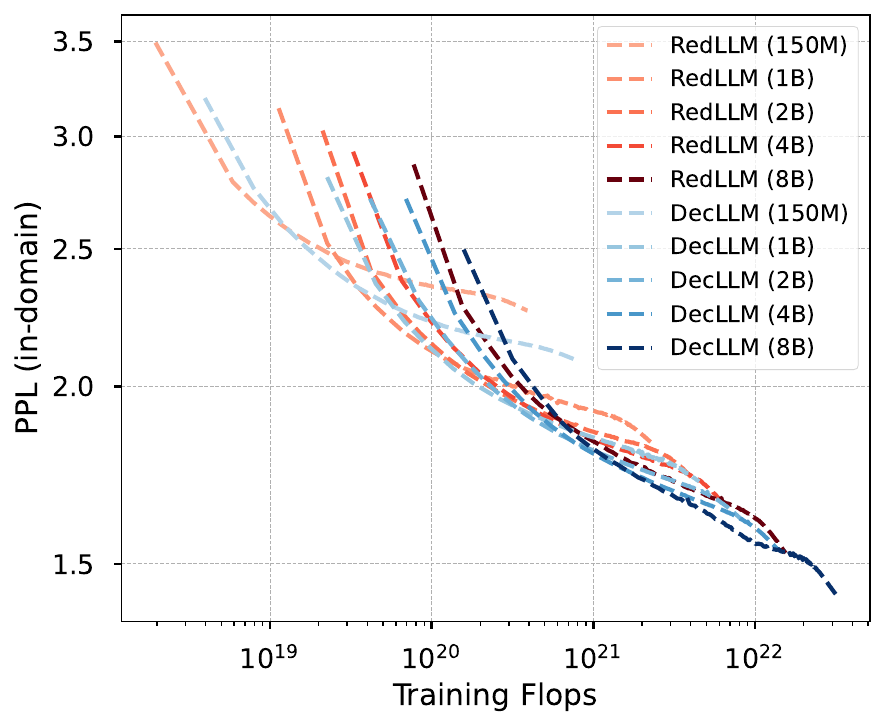}
    \includegraphics[width=0.40\textwidth]{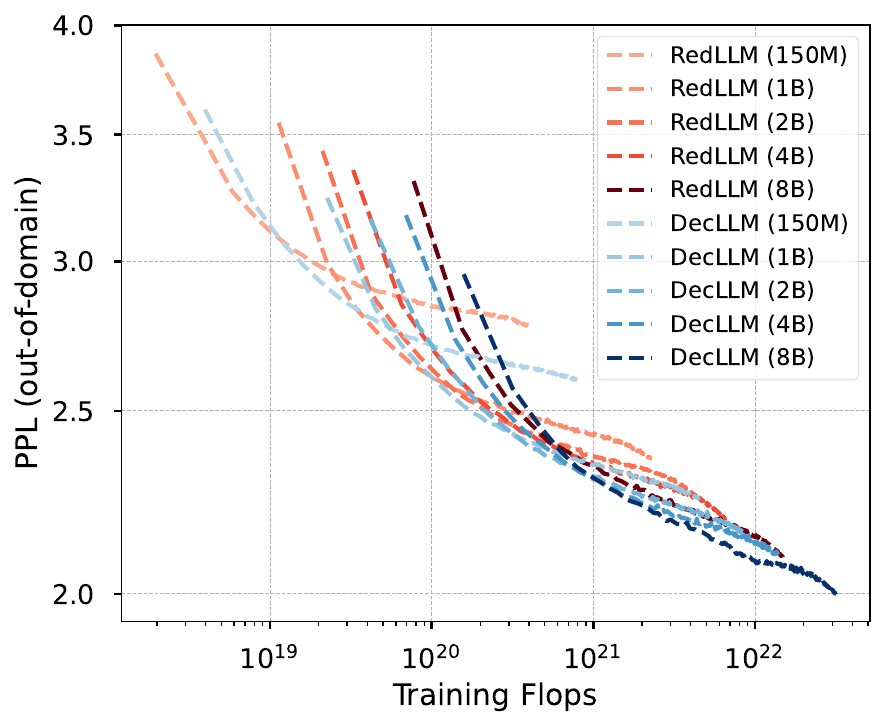}
    \caption{\label{fig:compute_optimal} PPL as a function of total training compute. Models are evaluated using the same prefix LM approach over different pretraining checkpoints.  The compute-optimal frontier is mostly dominated by \deconlyllm, especially with larger compute budget.}
\end{figure}

\paragraph{\redllm lags behind \deconlyllm for compute-optimal training.}

We next investigate how effectively different models utilize the compute, as measured by PPL during the pretraining. Intuitively, a model achieving lower PPL under the same compute budget is preferred. Figure \ref{fig:compute_optimal} and \ref{fig:compute_optimal_isoflops} show the comparison between \redllm and \deconlyllm with different compute budgets.
We first plot the PPL-to-compute curve for all models, where the bottom left region indicates the Pareto frontier. We then use this information to fit a compute-optimal scaling law function~\citep{hoffmann2022training} for both \redllm and \deconlyllm, and plot the isoFLOP for both. 
Figure \ref{fig:compute_optimal} (and \ref{fig:compute_optimal_isoflops}) shows that while \redllm shows a slight edge in low-compute settings, \deconlyllm's advantage becomes dominant as the compute scales up. This superior scaling behavior might come from its causal LM objective, which endows \deconlyllm with higher efficiency in utilizing training tokens. Still, the quality gap between \redllm and \deconlyllm is not that substantial and narrows with the increase of compute.

\begin{figure}[t]
    \small
    \centering
    \includegraphics[width=\textwidth]{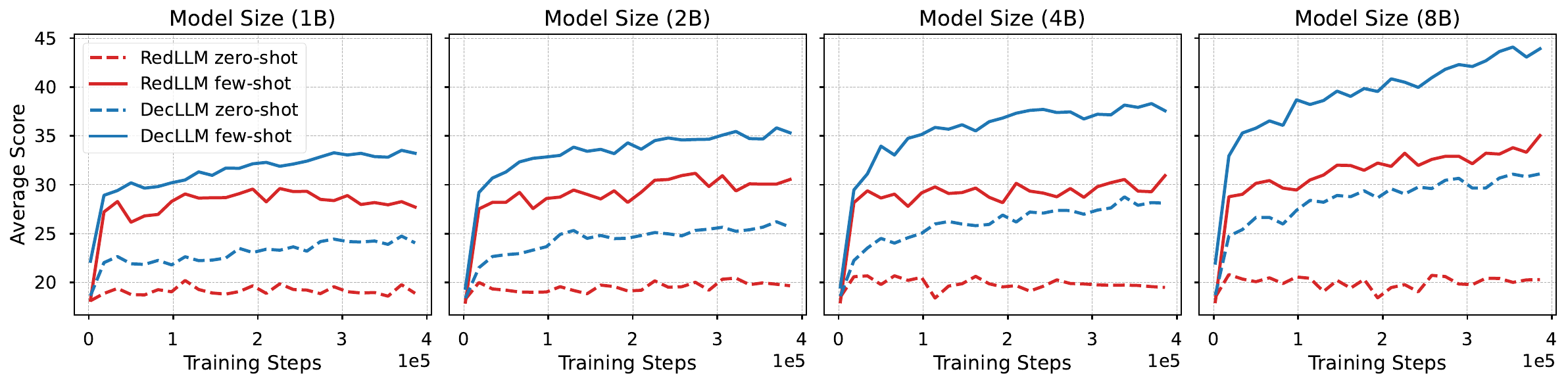}
    \caption{\label{fig:pretrain_perform_steps} Zero- and few-shot pretraining performance over training steps.}
\end{figure}

\paragraph{\redllm scales slightly for few-shot learning but is poor at zero-shot; both largely underperform \deconlyllm.}

While PPL has been widely used for measuring LLM capability, it may not be indicative enough for the downstream task-solving ability~\citep{kuribayashi-etal-2021-lower,hu2024can}. Apart from PPL, we further examine models' zero/few-shot performance as shown in Figure \ref{fig:pretrain_perform_steps}. \redllm and \deconlyllm both achieve non-trivial few-shot performance, demonstrating their capability of in-context learning. \deconlyllm yields consistently better performance as scaling up model and data size, for both zero-shot and few-shot learning. In contrast, \redllm performs badly at zero-shot learning, and only shows slight benefits from the scaling for few-shot learning. Its inferior zero-shot results, nevertheless, may not result from \redllm's insufficient capability but from an inadequate way of zero-shot prompting~\citep{patel2023bidirectional}. In general, \redllm falls behind \deconlyllm substantially at zero/few-shot learning during pretraining, resonating with the findings of~\citet{pmlr-v162-wang22u}. This still holds true even when their PPL scores are comparable, e.g., \redllm 8B vs. \deconlyllm 4B.

\begin{figure}[t]
    \small
    \centering
    \includegraphics[width=\textwidth]{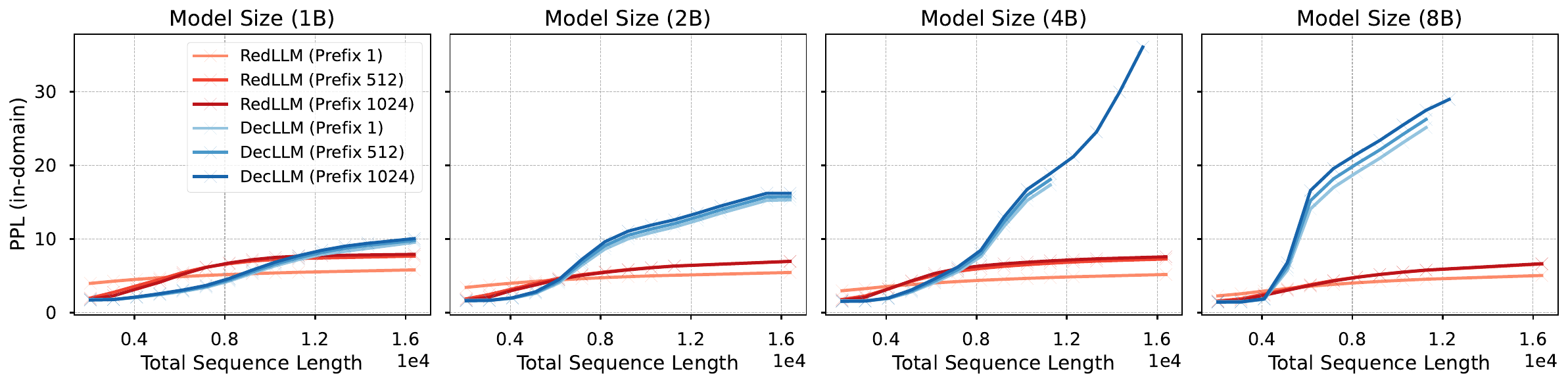}
    \caption{\label{fig:length_extra_indomain} PPL curves for length extrapolation on in-domain dataset (RedPajama). We follow the prefix LM evaluation approach and explore different prefix lengths (1, 512 and 1024).}
\end{figure}

\paragraph{\redllm shows comparable and even better length extrapolation capability than \deconlyllm along model scaling.}

One strength of \deconlyllm with rotary embedding is its length extrapolation ability, i.e., being able to handle sequences beyond pretraining context length. This raises the question of how \redllm performs on long sequence modeling and we show the results in Figure \ref{fig:length_extra_indomain} and \ref{fig:length_extrap_outdomain}. In general, PPL goes up -- becomes worse --  for both \redllm and \deconlyllm as context length increases, regardless of prefix lengths and domains.

\deconlyllm maintains relatively stable PPL up to a context length of 4096 tokens, i.e., twice the pretraining length. Beyond this point, PPL starts increasing, eventually exhibiting divergent behavior. We note that the larger \deconlyllm is, the more rapid PPL escalates as sequence length extends. In contrast, \redllm shows a surprisingly smoother increase in PPL, regardless of model sizes, demonstrating a stronger ability to extrapolate to longer sequences. Nevertheless, it often underperforms \deconlyllm on sequences shorter than 4K tokens.

Interestingly, the impact of prefix length is marginal on the extrapolation behavior except \redllm at prefix length of 1. \redllm suffers from significant quality drop on short sequences when reducing the prefix length to 1 since the capacity from its encoder becomes limited. As model scales up, however, the gap gets largely narrowed, perhaps due to the improved decoder capacity.

\begin{figure}[t]
    \small
    \centering
    \includegraphics[width=\textwidth]{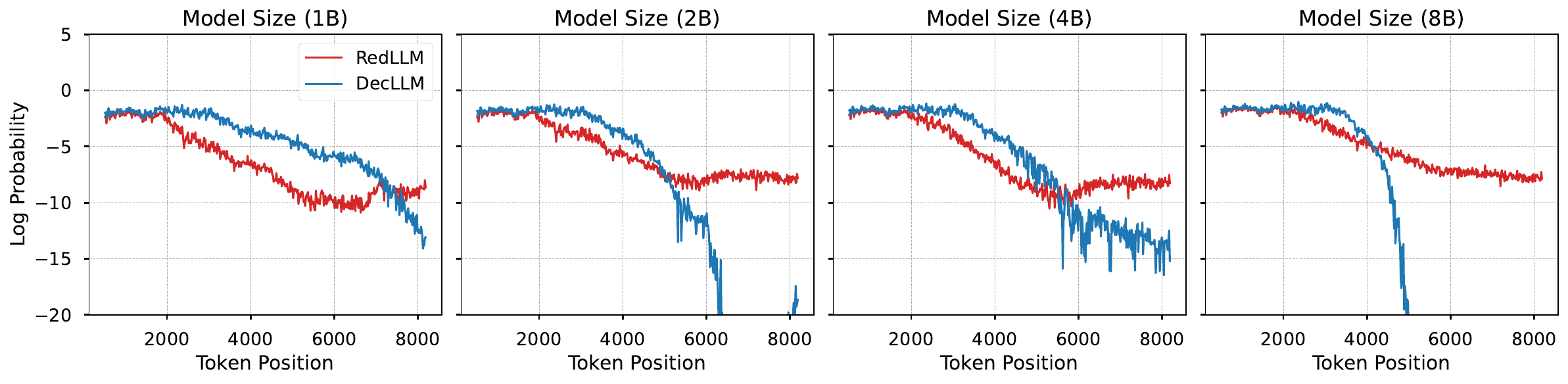}
    \includegraphics[width=\textwidth]{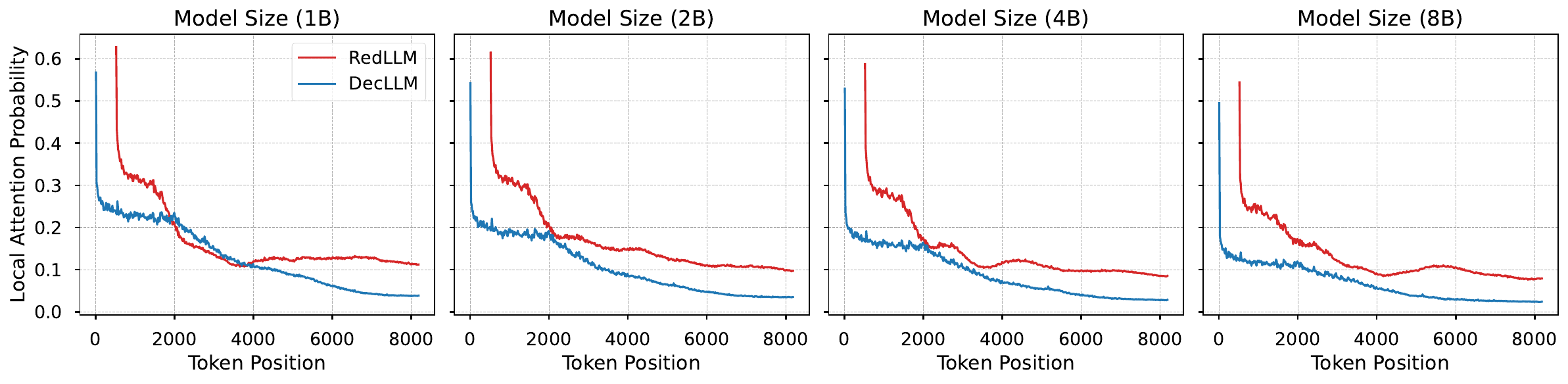}
    \caption{\label{fig:length_extra_analysis} Log probability of groundtruth target tokens (top) and local attention probability for decoder self-attention (bottom) as a function of token positions. We collect these statistics from 8 randomly sampled RedPajama examples, each with 512 prefix tokens and 7680 target tokens. We average the log probability for the same token position over different examples; for self-attention, we first average the attention weights over all heads, layers and examples, and then report the summed weights over a local window $[t-4, t]$ for token $x_t$.}
\end{figure}

\begin{figure}[t]
    \small
    \centering
    \begin{minipage}{\textwidth}
        \centering
        \small
        \subcaptionbox{\label{fig:encdec_can}\redllm: cross-attention.}{
            \includegraphics[width=0.31\textwidth]{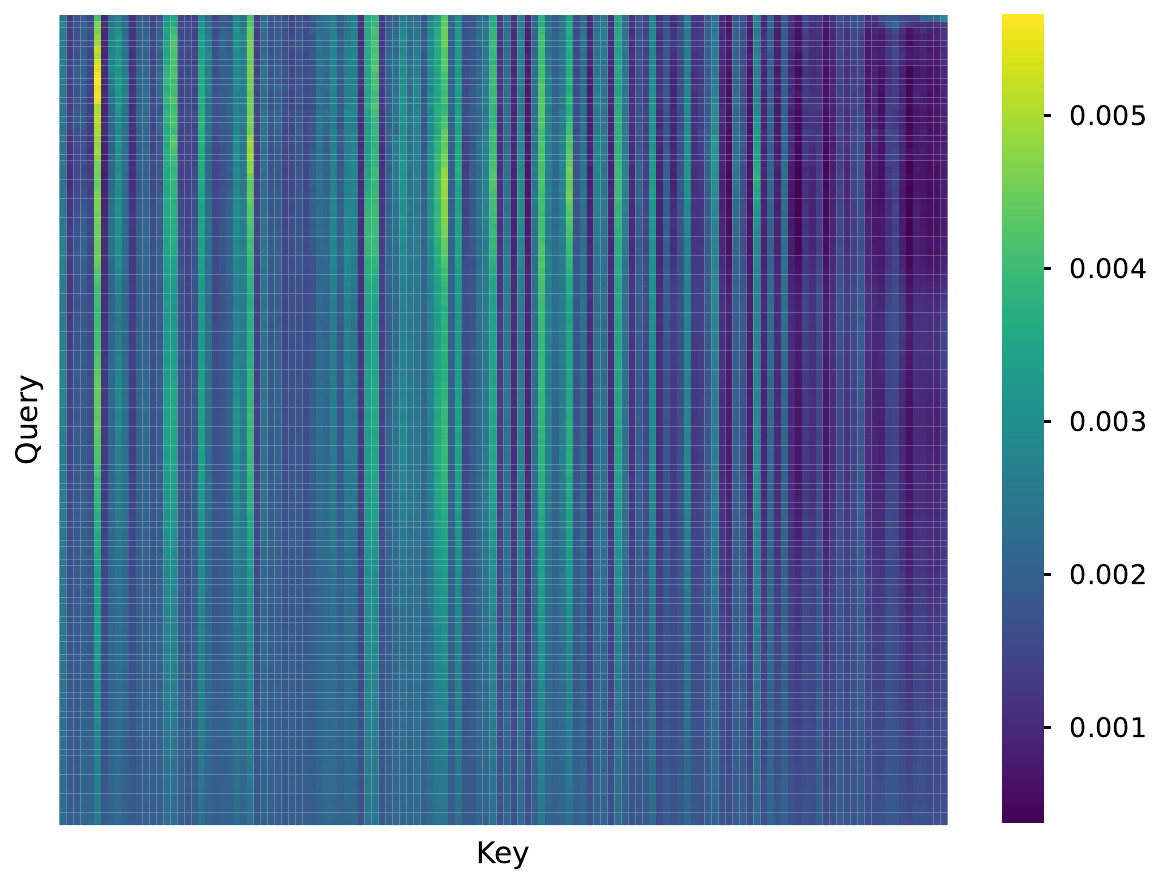}
        }
        \subcaptionbox{\label{fig:encdec_san}\redllm: self-attention.}{
            \includegraphics[width=0.31\textwidth]{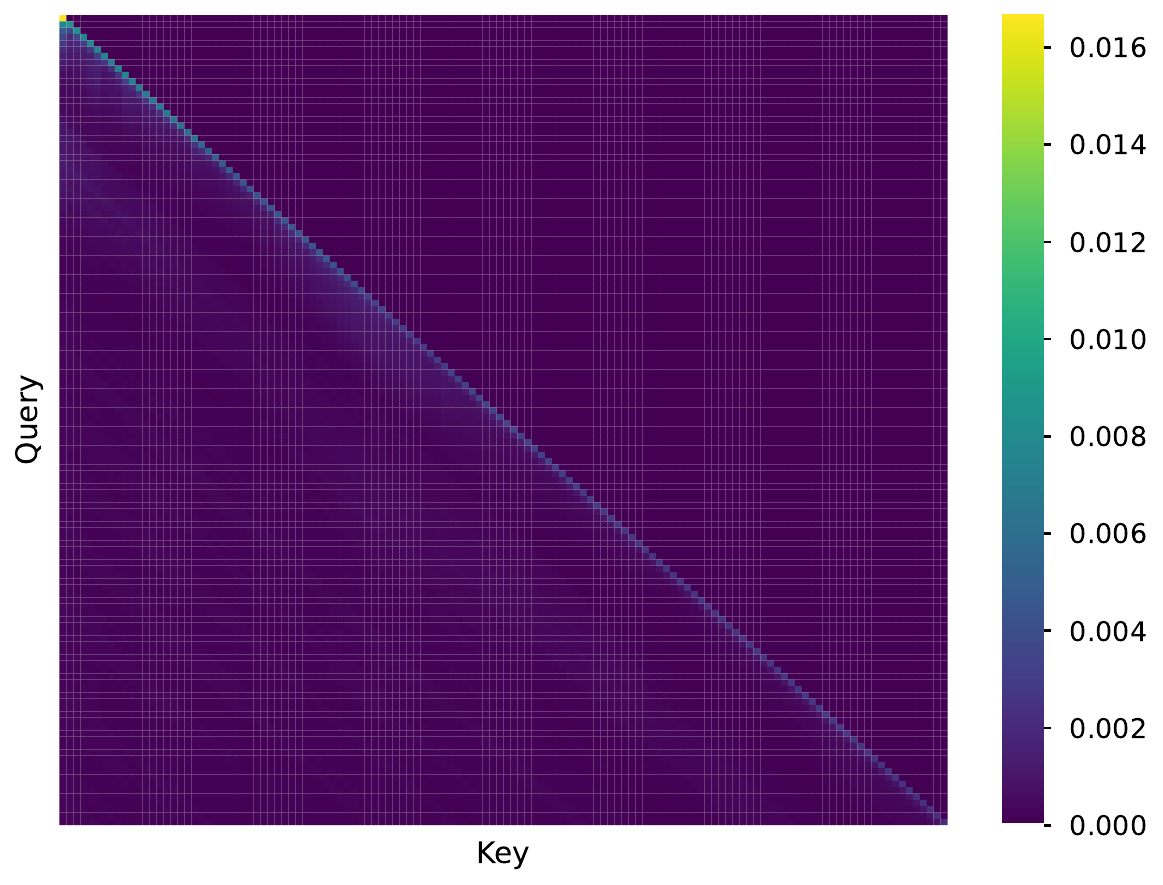}
        }
        \subcaptionbox{\label{fig:deconly_san}\deconlyllm: self-attention.}{
            \includegraphics[width=0.31\textwidth]{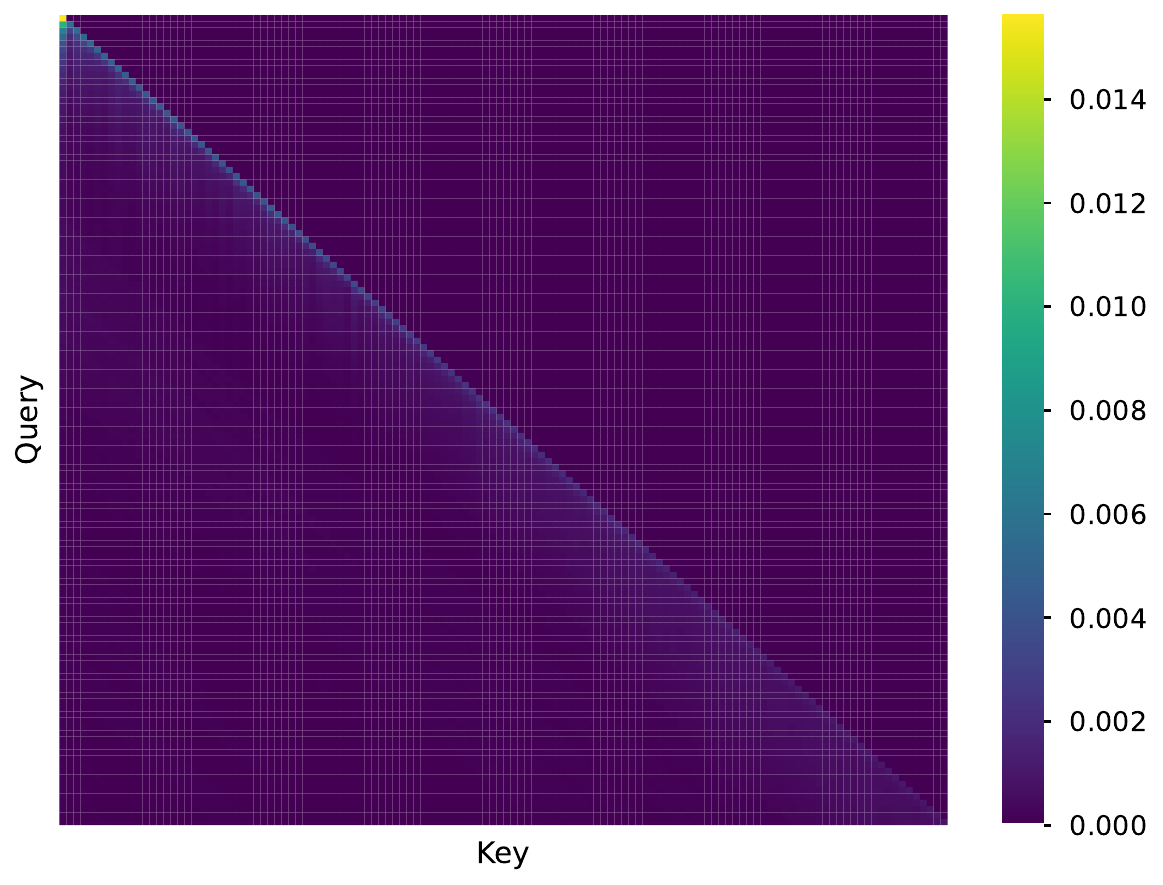}
        }
    \end{minipage}
    \caption{\label{fig:attn_weights_analysis} Attention weight visualization. x-axis: key; y-axis: query. The weights are first averaged over heads, layers and examples, and then compressed into a resolution of 128x128 through mean pooling with strides. Note self-attentions are from decoder layers. All models are 4B.}
\end{figure}

\paragraph{The decoder self- and cross-attention in \redllm show intriguing patterns under long context.}

The above results motivate us to dive deeper into the models for a better understanding of what happened under long context. We find from Figure \ref{fig:length_extra_analysis} (top) that the groundtruth target token's log probability often decreases as token position increases. In contrast to \redllm, \deconlyllm tends to suffer from a sharp probability drop after some point. We further inspect the decoder self-attention with the suspect that self-attention weights might not be functioning as intended. Figure \ref{fig:encdec_san} and \ref{fig:deconly_san} show that the decoder self-attention in \redllm and \deconlyllm follows a clear ``\textit{locality}'' structure where tokens favor attending to itself and nearby tokens, but the strength of locality weakens with the increase of token position particularly in \deconlyllm -- a phenomenon we called \textit{locality decay}. To verify this finding, we calculated the local attention probability, i.e., summed attention weights over a local window size of 5. Figure \ref{fig:length_extra_analysis} (bottom) shows that the decay behavior occurs in both \redllm and \deconlyllm, while \deconlyllm suffers more.

Figure \ref{fig:encdec_can} shows that the pattern of cross-attention weights differs significantly from that of self-attention weights. There is always a subset of input tokens attended by target tokens in \redllm, and this subset varies little~\citep{zhang2020sparsifying}. Such diverse attention may help \redllm to capture some distant information under long context modeling, but may also increase learning difficulty as the input-target dependency in language modeling is of higher uncertainty. Still, the underlying reason behind \redllm's long context behavior remains unclear, which we leave to the future.

\section{Finetuning Results}

\begin{figure}[t]
    \small
    \centering
    \includegraphics[width=\textwidth]{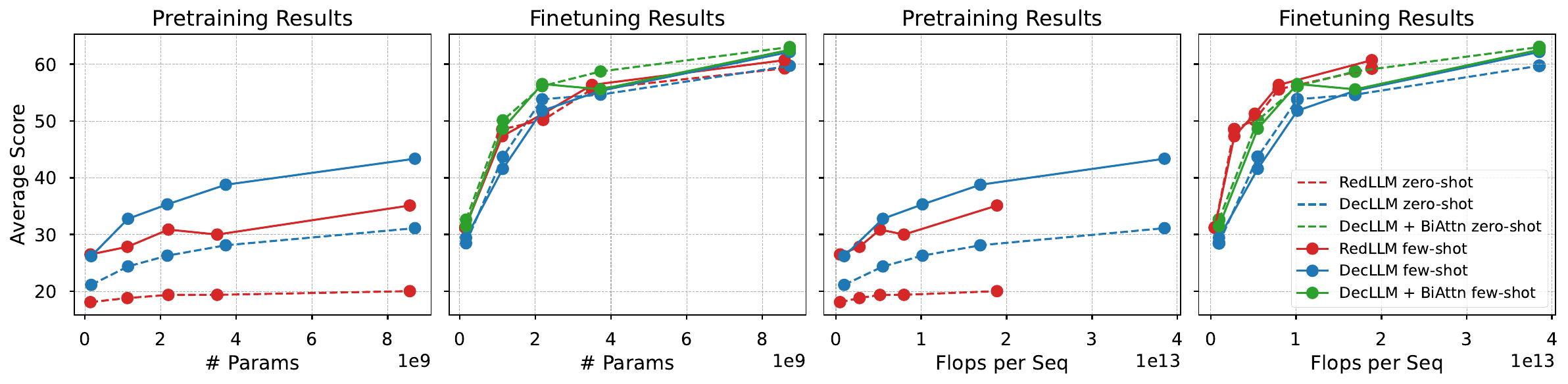}
    \caption{\label{fig:downstream_perform} Zero- and few-shot downstream performance as a function of model parameters (\#Params) and Flops. ``Flops per Seq'': inference Flops for a single sequence; ``\deconlyllm + BiAttn'': allowing bidirectional attention over inputs at finetuning.}
\end{figure}

\paragraph{\redllm shows high adaptability: matching and even surpassing \deconlyllm across scales after finetuning.}

While \redllm's zero/few-shot performance lags behind \deconlyllm in pretraining, finetuning changes the situation. Figure \ref{fig:downstream_perform} shows the averaged scaling results for finetuning along with pretraining under model parameters and inference Flops, and Table \ref{tab:downstream_perform} lists detailed numbers.
\redllm catches up with \deconlyllm after finetuning, even under the same amount of model parameters (where \redllm produces higher PPL than \deconlyllm). When switching the comparison basis to the inference Flops, \redllm shows significant advantage, nearly dominating the quality-compute Pareto frontier. The performance gap between zero-shot and few-shot evaluation is also largely reduced.
Apart from averaged results, we add per-task performance in Figures \ref{fig:downstream_pertask_1_6}, \ref{fig:downstream_pertask_7_12} and \ref{fig:downstream_pertask_13_16}. The results vary across tasks, e.g., \redllm largely surpasses \deconlyllm on ANLI, while \deconlyllm excels at ARC. These differences should be carefully considered when comparing both models.

Note \citet{pmlr-v162-wang22u} also observed similar results. However, their experiments were based on a 11B encoder-decoder model and 4.8B decoder-only model. In contrast, we present the scaling landscape across multiple model sizes. We further demonstrate the feasibility of achieving comparable performance to \deconlyllm under similar model sizes.

In addition, these results suggest that the adaptability of different models varies greatly. Models with worse pretraining performance may stand out after finetuning, such as \redllm. We conjecture that the superior pretraining performance of \deconlyllm is mostly caused by the higher degree of matching between its pretraining objective and the downstream evaluation, rather than its stronger capability. Surprisingly, PPL also fails to capture a model's adaptability, echoing with~\citet{hu2024can}.

\paragraph{Bidirectional input attention improves \deconlyllm greatly but doesn't change the quality-compute frontier.}

An important difference between \redllm and \deconlyllm at finetuning is the bidirectional input attention (BiAttn) in \redllm's encoder, which allows input tokens attending to each other. Such a structure often improves input understanding, and has proven useful in various settings including \deconlyllm~\citep{pmlr-v162-wang22u,pmlr-v162-zhang22h}. This may explain why \deconlyllm underperforms \redllm after finetuning.

We thus enable this structure for \deconlyllm. Figure \ref{fig:downstream_perform} and Table \ref{tab:downstream_perform} show that the performance of \deconlyllm gets substantially improved. Interestingly, while \deconlyllm + BiAttn largely outperforms \redllm at zero-shot learning, their few-shot quality gap becomes smaller to negligible. Still, the overall quality gap among different models is much smaller compared to the pretraining performance gap, and the leading position of \redllm with respect to the quality-compute trade-off remains.

\begin{figure}[t]
    \small
    \centering
    \includegraphics[width=0.7\textwidth]{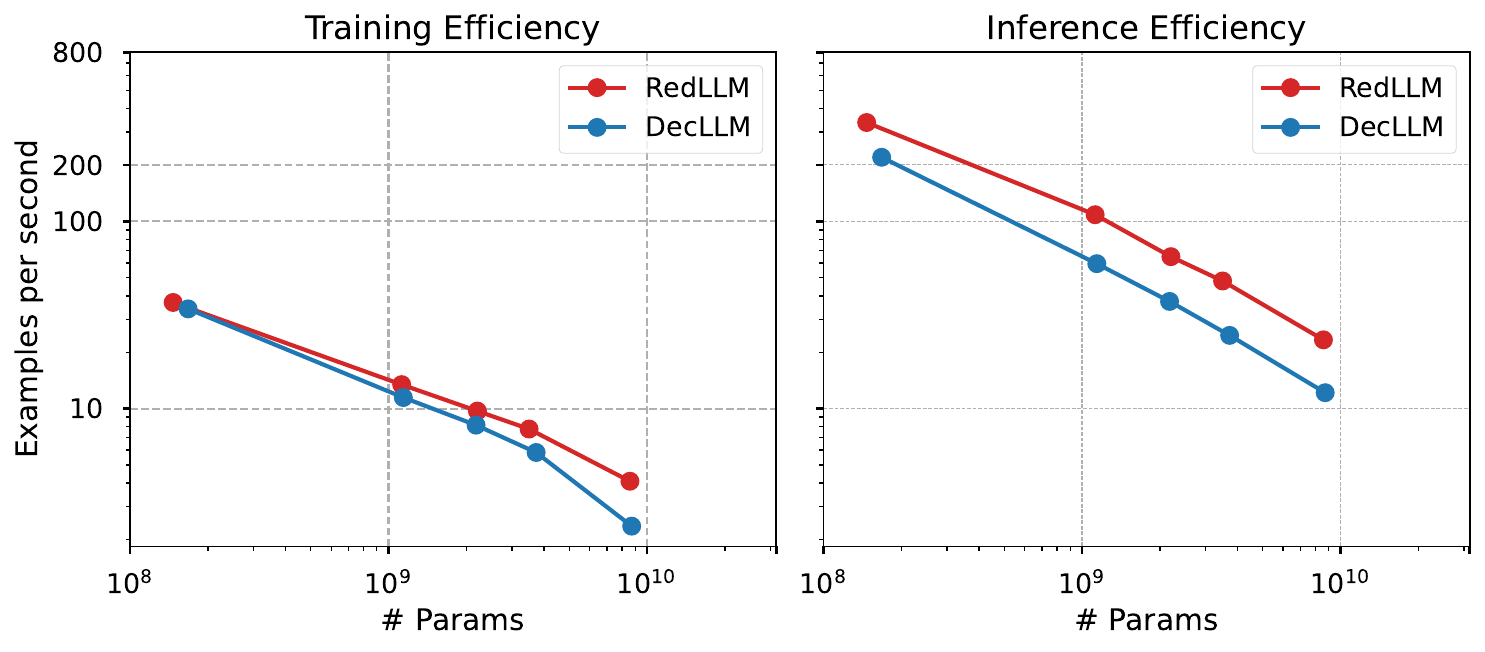}
    \caption{\label{fig:efficiency} Efficiency analysis. We show the number of examples processed per second for efficiency comparison. Results are collected with batch size of 1 using 2 TPU v5p chips.}
\end{figure}

\paragraph{\redllm has clear advantage over \deconlyllm on training and inference efficiency.}

Beyond theoretical Flops analysis, our empirical efficiency analysis (Figure \ref{fig:efficiency}) reveals a striking advantage for \redllm in both training and inference throughput. This advantage becomes even more pronounced as model size increases, particularly during pretraining, where \redllm significantly outpaces \deconlyllm.
This remarkable efficiency, together with \redllm's promising finetuning performance, underscores its significance, particularly on the fact that instruction tuning and alignment have become an essential part of LLM development~\citep{ouyang2022training}.

\section{Conclusion and Future Work}

Encoder-decoder or decoder-only for LLM? We answered this question from the scaling perspective and provided a comprehensive analysis over multiple dimensions. We conducted large-scale pretraining and finetuning experiments using RedPajama V1 and FLAN, with models ranging from 150M to 8B parameters. Our results show distinct strengths for each architecture. \deconlyllm presents unique strengths during pretraining, including compute-optimal scaling, zero/few-shot learning, and strong training stability. By contrast, \redllm excels in finetuning scenarios, showing superior running efficiency and finetuning performance. Besides, \redllm demonstrates comparable scaling properties and context length extrapolation capability.

There are many potential future directions. We are particularly interested in exploring the scalability of \redllm beyond 8B parameters. Our current study focused on balanced architectures, where the encoder and decoder have an equal number of layers. Investigating imbalanced architectures, such as those with a deep encoder and a shallow decoder, presents a compelling avenue. Besides, a comparative analysis of various pretraining objectives and a deeper understanding of how \redllm extrapolates to longer sequences would be valuable.

\subsubsection*{Acknowledgments}

We’d like to thank Tianqi Liu, Tris Warkentin, and Noam Shazeer for their constructive feedback on the manuscript. We also thank the T5X team~\citep{roberts2023scaling} for the training framework.

\bibliography{paper}
\bibliographystyle{iclr2025_conference}

\clearpage
\appendix
\section{Appendix}

\begin{figure}[!ht]
    \small
    \centering
    \includegraphics[width=0.7\textwidth]{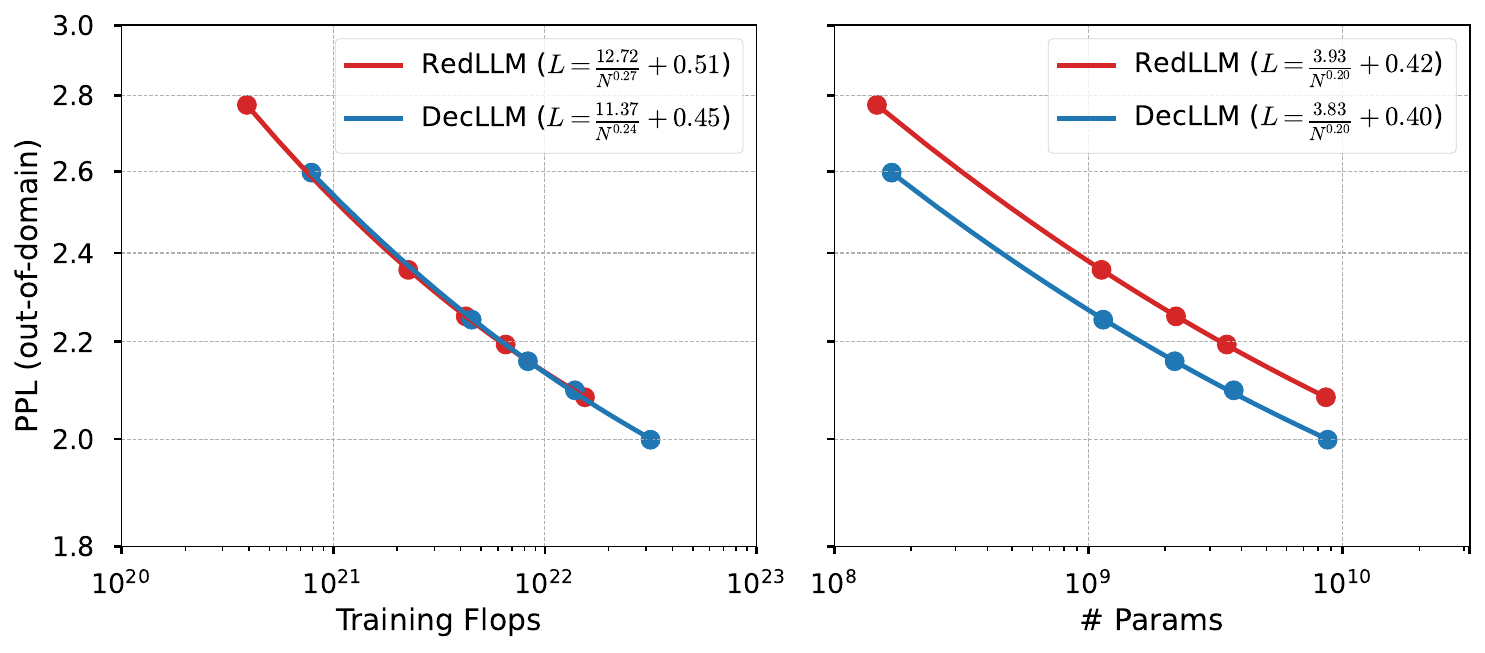}
    \caption{\label{fig:scaling_law_outdomain} Fitted scaling law on out-of-domain dataset (Paloma). The scaling behavior is consistent with the in-domain evaluation.}
\end{figure}

\begin{figure}[!ht]
\centering
\includegraphics[width=0.8\textwidth]{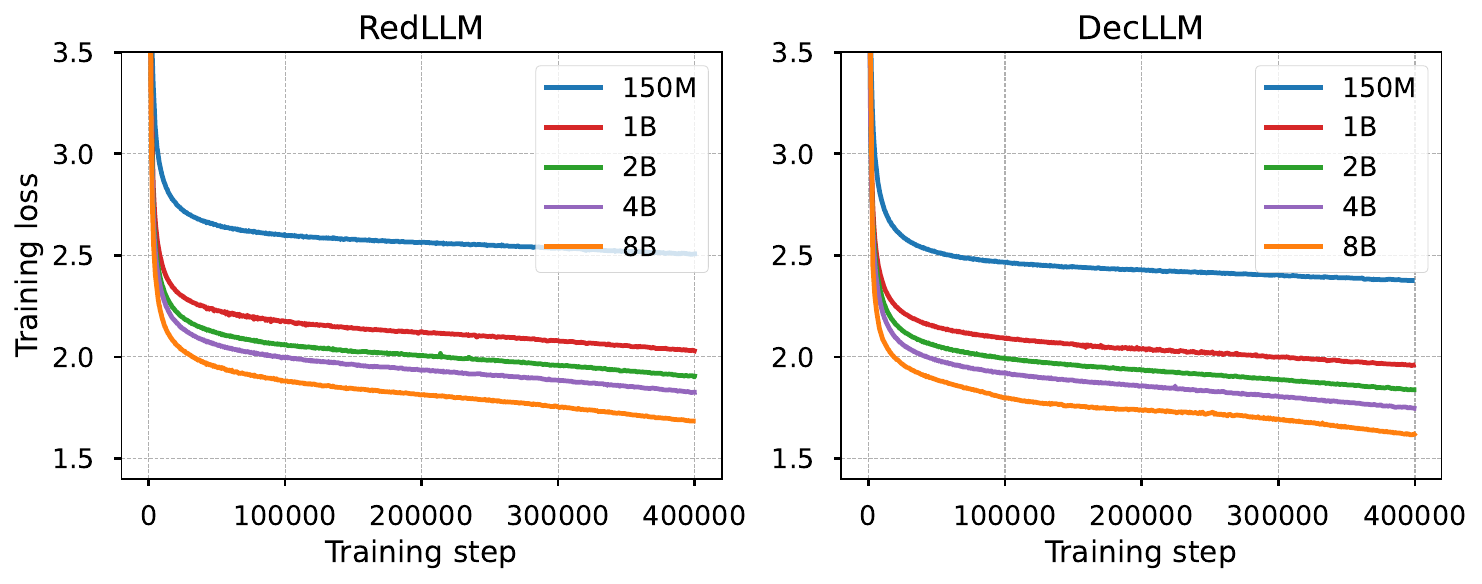}
\caption{\label{fig:training_loss} Pretraining loss curve. Note the loss is not directly comparable between \redllm and \deconlyllm due to their difference in pretraining objective.}
\end{figure}

\begin{figure}[!ht]
    \small
    \centering
    \includegraphics[width=\textwidth]{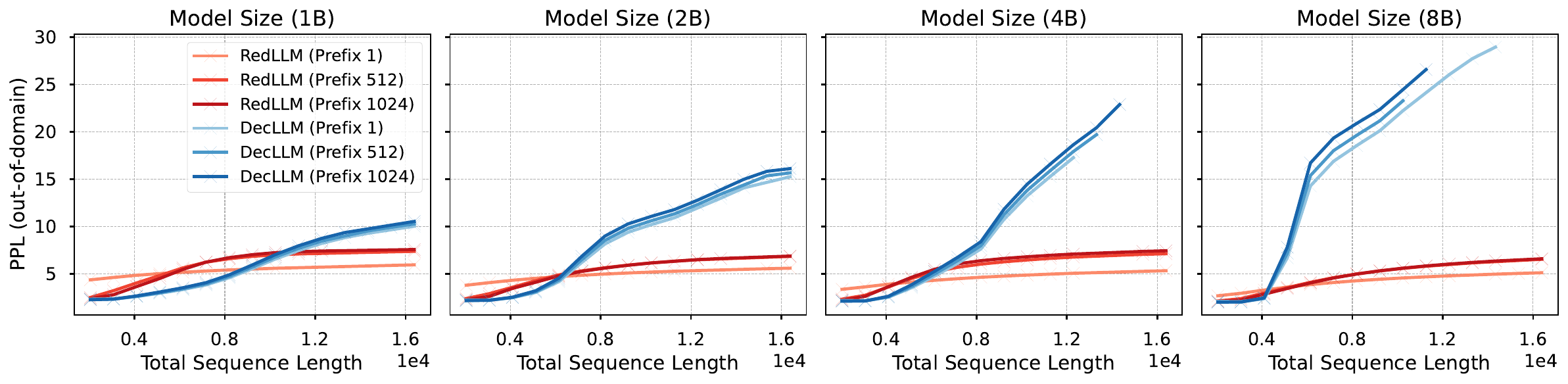}
    \caption{\label{fig:length_extrap_outdomain} PPL curves for length extrapolation on out-of-domain dataset (Paloma). The extrapolation result is consistent with the in-domain evaluation.}
\end{figure}

\begin{figure}[t]
    \small
    \centering
        \includegraphics[width=0.40\textwidth]{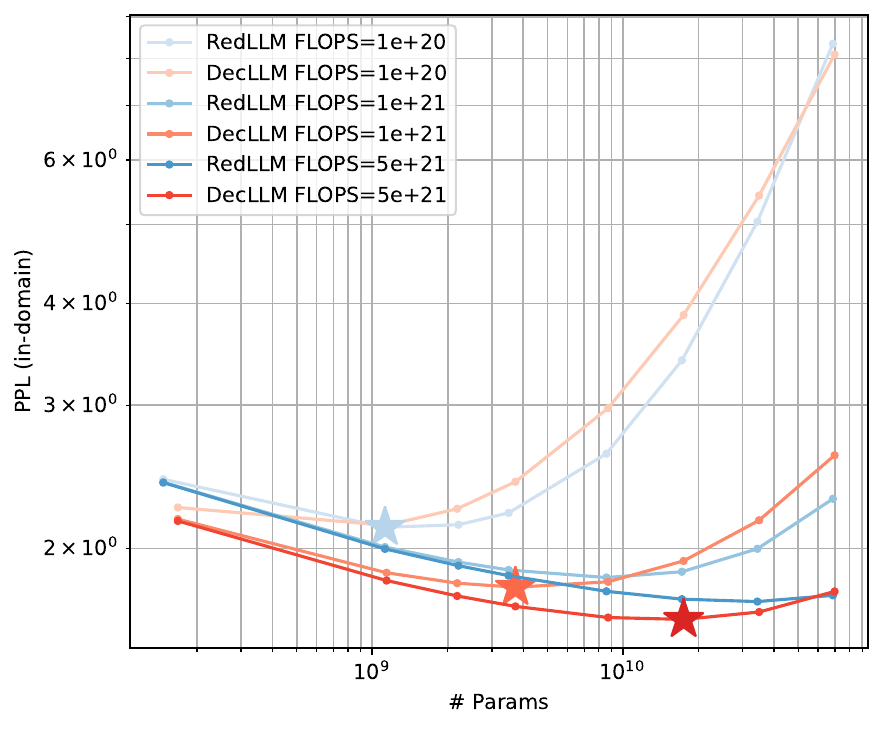}
    \includegraphics[width=0.40\textwidth]{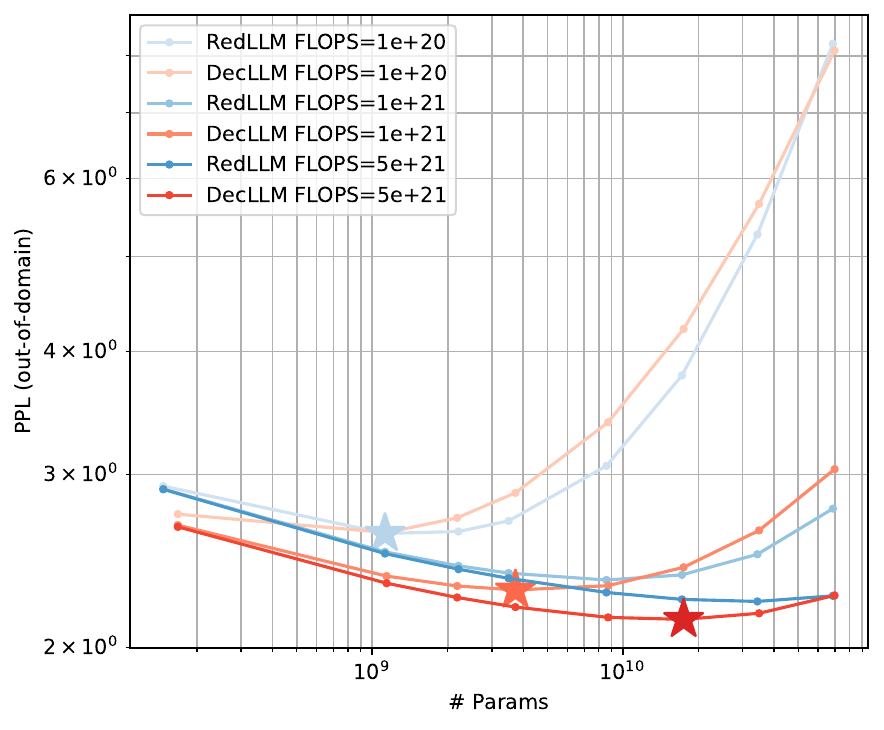}
    \caption{\label{fig:compute_optimal_isoflops} IsoFLOP slices of \redllm and \deconlyllm with three different compute budget. $\star$ indicates the compute-optimal configuration for each compute budget. Note \redllm curves are plotted in blue.}
\end{figure}

\begin{figure}[!ht]
\centering
\includegraphics[width=\textwidth]{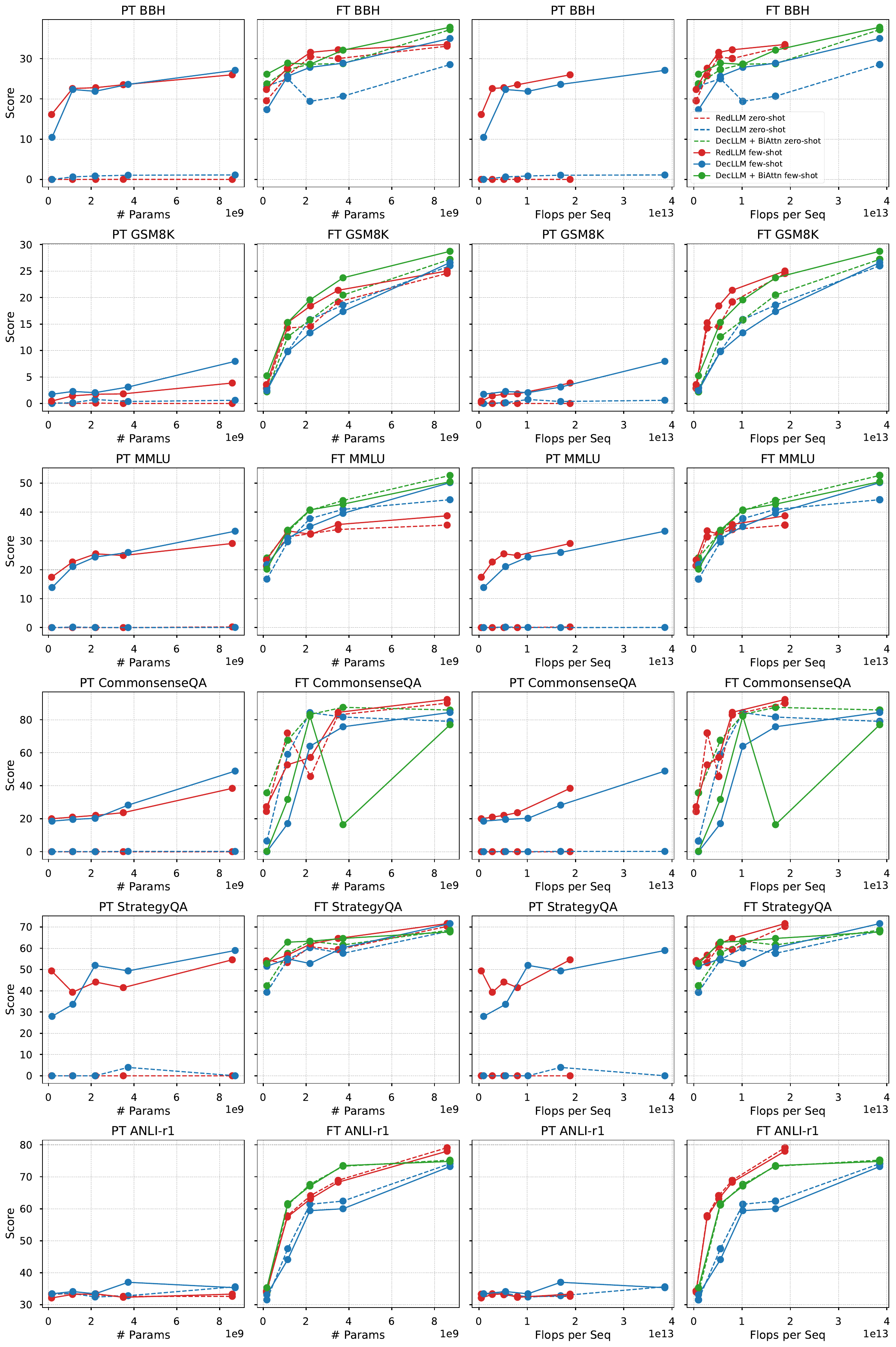}
\caption{\label{fig:downstream_pertask_1_6} Zero- and few-shot downstream performance per task. ``PT'': pretraining; ``FT'': finetuning.}
\end{figure}

\begin{figure}[!ht]
\centering
\includegraphics[width=\textwidth]{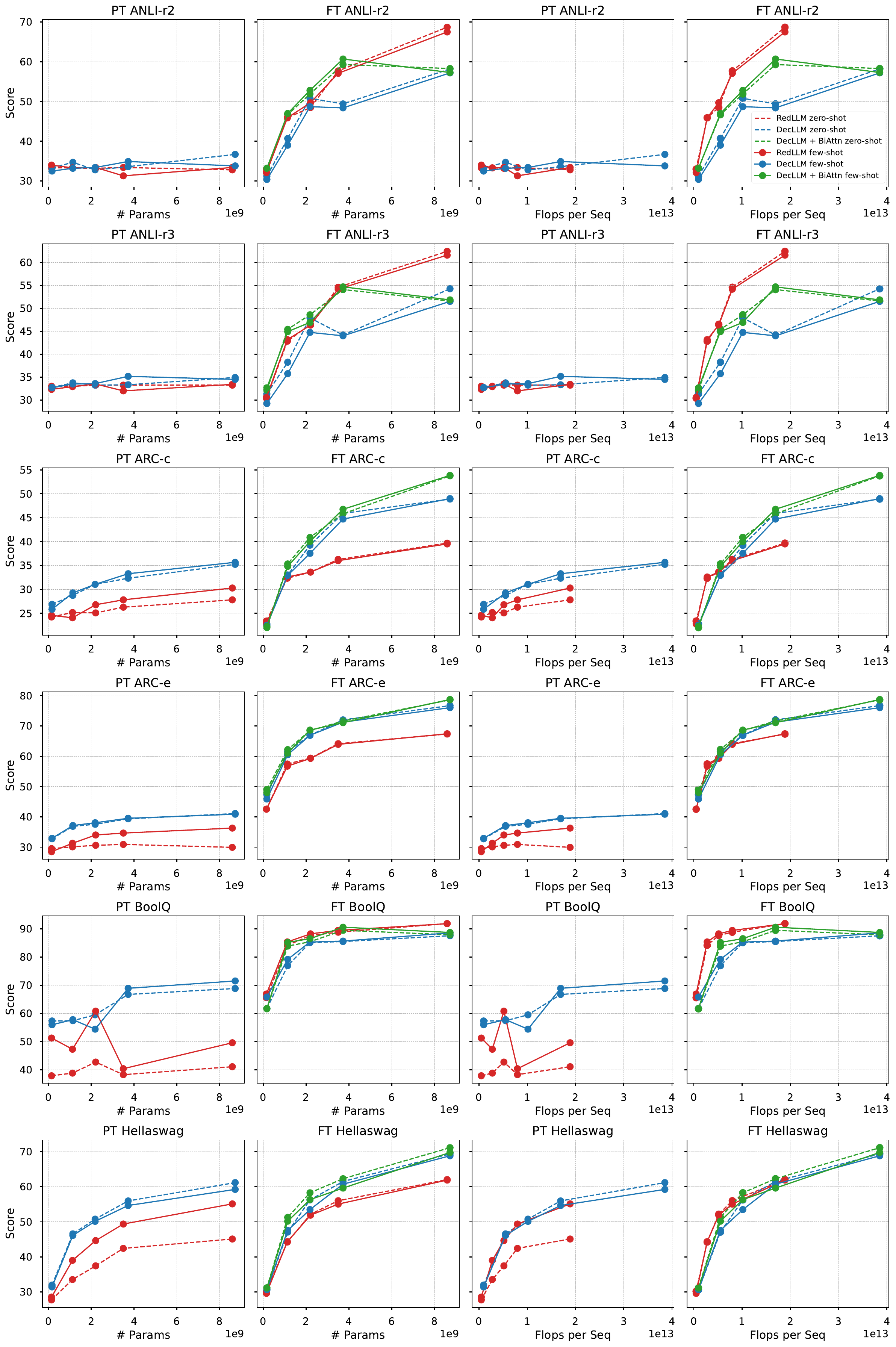}
\caption{\label{fig:downstream_pertask_7_12} Zero- and few-shot downstream performance per task. ``PT'': pretraining; ``FT'': finetuning.}
\end{figure}

\begin{figure}[!ht]
\centering
\includegraphics[width=\textwidth]{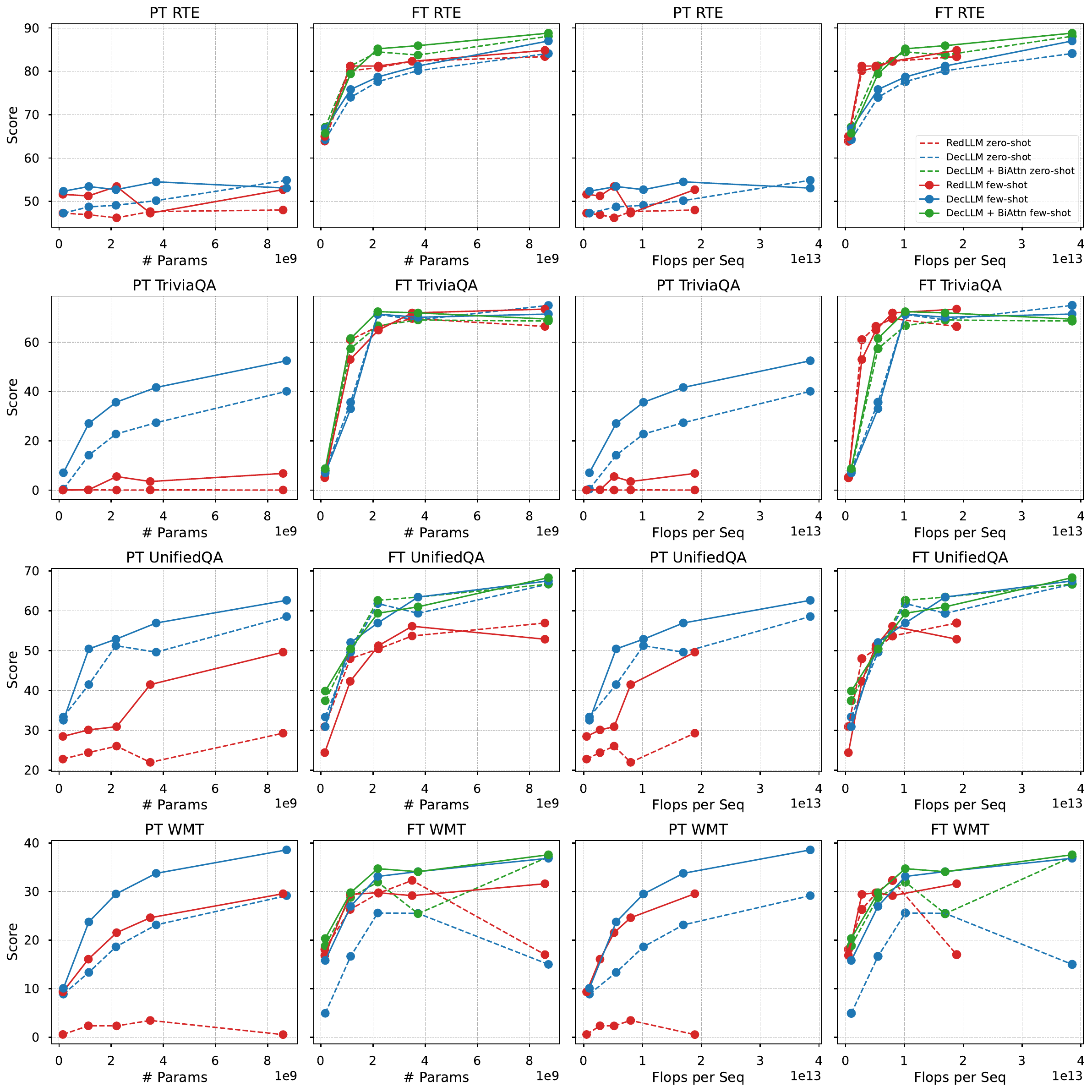}
\caption{\label{fig:downstream_pertask_13_16} Zero- and few-shot downstream performance per task. ``PT'': pretraining; ``FT'': finetuning.}
\end{figure}

\begin{table}[t]
\centering
\centering
\small
\begin{tabular}{lllrrrrr}
\toprule
\multicolumn{3}{c}{Setup} & 150M & 1B & 2B & 4B & 8B \\
\midrule
\multirow{4}{*}{Pretraining}
& \multirow{2}{*}{Zero-Shot}
    & \redllm & 18.11 & 18.82 & 19.38 & 19.39 & 20.04 \\
&   & \deconlyllm & 21.14 & 24.39 & 26.29 & 28.12 & 31.13 \\
\cmidrule(lr){2-8}
& \multirow{2}{*}{Few-Shot}
    & \redllm & 26.51 & 27.84 & 30.88 & 30.01 & 35.13 \\
&   & \deconlyllm & 26.21 & 32.79 & 35.33 & 38.79 & 43.37 \\
\midrule
\multirow{6}{*}{Finetuning}
& \multirow{3}{*}{Zero-Shot} 
    & \redllm & 31.23 & 48.55 & 50.19 & 55.61 & 59.69 \\
&   & \deconlyllm & 29.97 & 43.70 & 53.84 & 54.63 & 58.26 \\
&   & \quad + BiAttn & 33.73 & 50.12 & 56.15 & 58.07 & 63.03 \\
\cmidrule(lr){2-8}
& \multirow{3}{*}{Few-Shot}
    & \redllm & 31.24 & 47.32 & 51.30 & 56.37 & 61.32 \\
&   & \deconlyllm & 30.14 & 41.58 & 51.82 & 57.22 & 59.02 \\
&   & \quad + BiAttn & 31.50 & 48.13 & 56.52 & 55.95 & 62.54 \\
\bottomrule
\end{tabular}

\caption{\label{tab:downstream_perform} Results for zero- and few-shot downstream performance. ``+ BiAttn'': \deconlyllm with bidirectional input attention enabled. }
\end{table}

\end{document}